\documentclass[11pt,a4paper]{article}
\usepackage[hyperref]{acl2020}
\usepackage{times}
\usepackage{latexsym}

\usepackage{microtype}

\usepackage{multirow}
\usepackage{graphicx}
\usepackage{float}

\aclfinalcopy 
\def\aclpaperid{3084} 

\newcommand\s{\mbox{/-s/}}
\newcommand\en{\mbox{/-(e)n/}}
\newcommand\er{\mbox{/-er/}}
\newcommand\e{\mbox{/-e/}}
\newcommand\z{\mbox{/-$\emptyset$/}}

\title{Inflecting when there's no majority: Limitations of encoder-decoder neural networks as cognitive models for German plurals}

\author{Kate McCurdy \quad Sharon Goldwater \quad Adam Lopez \\
  Institute for Language, Cognition and Computation\\
  School of Informatics \\ University of Edinburgh \\
  \tt{kate.mccurdy@ed.ac.uk, \{sgwater, alopez\}@inf.ed.ac.uk}}

\date{}

\begin{document}
\maketitle
\begin{abstract}
Can artificial neural networks learn to represent inflectional morphology and generalize to new words as human speakers do? 
\citet{kirovRecurrentNeuralNetworks2018} argue that the answer is yes: modern Encoder-Decoder (ED) architectures learn human-like behavior when inflecting English verbs, such as extending the regular past tense form \mbox{/-(e)d/} to novel words. However, their work does not address the criticism raised by \citet{marcusGermanInflectionException1995}: that neural models may learn to extend not the \textit{regular}, but the \textit{most frequent} class --- and thus fail on tasks like German number inflection, where infrequent suffixes like \s\ can still be productively generalized.
To investigate this question, we first collect a new dataset from German speakers (production and ratings of plural forms for novel nouns) that is designed to avoid sources of information unavailable to the ED model.
The speaker data show high variability, and two suffixes evince `regular' behavior, appearing more often with phonologically atypical inputs. Encoder-decoder models do generalize the most frequently produced plural class, but do not show human-like variability or `regular' extension of these other plural markers. We conclude that modern neural models may still struggle with minority-class generalization.
\end{abstract}

\section{Introduction}

Morphology has historically been the site of vigorous debate on the capacity of neural models to capture human speaker behavior, and hence ground claims about speaker cognition. In 1986, \citeauthor{rumelhartLearningTensesEnglish1986} described a neural network model which learned to map English present tense verbs to their past tense forms. Importantly, the network handled both \textbf{regular} verbs, whose past tense is formed systematically by adding the suffix \mbox{/-(e)d/} (e.g.\ \textit{jumped}), and \textbf{irregular} verbs where the 
present and past tenses bear no systematic relationship (e.g.\ \textit{ran}). The authors suggested their model provided ``an alternative [...] to the implicit knowledge of rules" (\citeyear[][218]{rumelhartLearningTensesEnglish1986}), a claim which sparked considerable controversy. 
\citet{pinkerLanguageConnectionismAnalysis1988} highlighted many empirical inadequacies of the \citeauthor{rumelhartLearningTensesEnglish1986} model, and argued that these failures stemmed from ``central features of connectionist ideology" and would persist in any neural network model lacking a symbolic processing component. 

Recently, however, \citet[][henceforth K\&C]{kirovRecurrentNeuralNetworks2018}  revisited the English past tense debate and showed that modern recurrent neural networks with encoder-decoder (ED) architectures overcome many of the empirical limitations of earlier neural models. Their ED model successfully learns to generalize the regular past tense suffix \mbox{/-(e)d/}, achieving near-ceiling accuracy on held-out test data. Moreover, its errors result from overapplication of the regular past tense (e.g. \textit{throw--throwed})---a type of error observed in human language learners as well---as opposed to the unattested forms produced by \citeauthor{rumelhartLearningTensesEnglish1986}'s model. K\&C conclude that modern neural networks can learn human-like behavior for English past tense without recourse to explicit symbolic structure, and invite researchers to move beyond the `rules' debate, asking instead whether the learner correctly generalizes to a range of novel inputs, and whether its errors (and other behavior) are human-like.

This challenge was first taken up by \citet{corkeryAreWeThere2019}, who showed that, on novel English-like words designed to elicit some irregular generalizations from humans, the ED model's predictions do not closely match the human data. While these results suggest possible problems with the ED model, English may not be the best test case to fully understand these, since the sole regular inflectional class is also by far the most frequent.
In contrast, many languages have multiple inflectional classes which can act `regular' under various conditions \citep{seidenbergQuasiregularityItsDiscontents2014, clahsenContributionsLinguisticTypology2016}. 

In this paper, we examine
German number inflection, which has been identified as a crucial test case for connectionist modeling \citep{kopckeSchemasGermanPlural1988, bybeeRegularMorphologyLexicon1995, marcusGermanInflectionException1995, clahsenLexicalEntriesRules1999}. The German plural system features eight plural markers (c.f. Table \ref{table:de_plurals}), none of which hold a numerical majority in type or token frequency.  Different linguistic environments favor different plural markers \citep[e.g.][]{kopckeSchemasGermanPlural1988, wiesePhonologyGerman1996, yangPriceLinguisticProductivity2016}, and even the famously rare suffix \s\ is nonetheless \textbf{productive}, in the sense that speakers readily extend it to new words.\footnote{For example, the Institut f\"ur Deutsche Sprache (\url{https://www.owid.de/service/stichwortlisten/neo_neuste}) officially added multiple \s-inflecting nouns to the German language in 2019, including \textit{Verh\"utungsapp, Morphsuit} and \textit{Onesie}.
} 
In their analysis of the German plural system, \citet[][henceforth M95]{marcusGermanInflectionException1995} argue that neural networks generalize the most frequent patterns to unfamiliar inputs, and thus
struggle to represent productive but rare classes such as \s.
We investigate that claim using the novel German-like nouns M95 developed. 

Because the design and results of previous human studies have been somewhat inconsistent, and because we want to compare to fine-grained results from individuals (not just published averages), we first collect a new dataset of plural productions and ratings from German speakers. 
Our speaker data show high variability: no class holds a majority overall, and two less frequent suffixes show a relative preference for phonologically atypical inputs (``Non-Rhymes").
We then compare our human data with the predictions of the encoder-decoder (ED) model proposed by K\&C. 
While our human data paint a more complex picture of the German plural system than M95 claimed, nevertheless M95's central idea is borne out: when given Non-Rhymes, the ED model prefers the most frequent plural class, but speakers behave differently.
This finding reveals that while modern neural models are far more powerful than earlier ones, they still have limitations as models of cognition in contexts like German number inflection, where no class holds a majority.
The model may correctly identify the most frequent class, but fails to learn the conditions under which minority classes are productive for speakers.

\section{Study 1: Speaker plural inflection}

To evaluate whether neural models generalize correctly, we need to compare their behavior with that of humans on the same task. Unfortunately, no existing datasets were suitable, so our first study
asks how German speakers inflect novel nouns.

\subsection{Background}

\paragraph{Wug testing and productivity} 
If an English speaker needs to produce the plural form of an unknown word such as \textit{wug}, that speaker must decide whether \textit{wug} belongs to the same inflectional class as \textit{dog} and \textit{cat} (yielding plural \textit{wugs}) or the same class as \textit{sheep} and \textit{deer} (yielding \textit{wug}). Speakers' overwhelming preference for \textit{wugs} in this scenario indicates that the \s\ plural class is productive in English: a productive morphological process can be generalized to new inputs. This task of inflecting novel (\textbf{nonce}) words is known as the \textit{wug test} \citep{berkoChildLearningEnglish1958}, and is the standard method to determine productivity in psycholinguistic research.
While the concept of morphological `regularity' is not well-defined \citep{herceDeconstructingIrRegularity2019}, productivity is nonetheless an essential component: an inflectional class that is not productive cannot be regular.

\begin{table}[]
\begin{center}
 \begin{tabular}{c c c c c} 
 Suffix & Singular & Plural & Type & Token \\ 
 \hline\hline
 /-(e)n/ & Strasse & Strassen & 48\% & 45\% \\ 
 \hline
  \multirow{2}{2em}{/-e/} & Hund & Hunde &  27\% & 21\% \\
 & Kuh & K\"uhe &  &  \\
 \hline
  \multirow{2}{2em}{\z} & Daumen & Daumen &  17\% & 29\% \\
 & Mutter & M\"utter &  &  \\
 \hline
  \multirow{2}{2em}{/-er/} & Kind & Kinder & 4\% & 3\% \\
 & Wald & W\"alder &  & \\
 \hline
 /-s/ & Auto & Autos & 4\% & 2\% \\
\end{tabular}
\caption{German plural system with examples, ordered by CELEX type frequency \citep{sonnenstuhlProcessingRepresentationGerman2002}.} 
\label{table:de_plurals}
\end{center}
\end{table}

\paragraph{Productivity in German plurals}  The German plural system comprises five suffixes: \e, \er, \z\footnote{\z\ refers to the so-called ``zero plural", and is indicated as ``zero" on all figures in this paper.}, \en, and \s. The first three can optionally combine with an umlaut over the root vowel.\footnote{ Umlaut is a process which fronts a back vowel, so only roots with back vowels can take an umlaut (e.g. \textit{Dach $\rightarrow$ D\"acher}, \textit{Fuss $\rightarrow$ F\"usse}).} Umlaut varies semi-independently of plural class \citep{wiesePhonologyGerman1996}, and is not fully predictable; for simplicity, this study will focus only on the five main suffix classes for analysis.
Examples in all forms are shown in Table \ref{table:de_plurals}. Each plural suffix is also shown with its \textbf{type frequency} (counting each word type only once, how many types in the lexicon take this plural?)\ and \textbf{token frequency} (how often do words with this plural suffix appear in the corpus overall?). German nouns can have one of three grammatical genders --- masculine, feminine, or neuter --- and this lexical feature is highly associated with plural class: most feminine nouns take \en, while \e\ and \z\ nouns are often masculine or neuter. The phonological shape of a noun also influences its plural class; for example, most nouns ending with schwa take \en\ \citep{elsenAcquisitionGermanPlurals2002}.
Although there are statistical tendencies, there are no absolute rules, and no suffix holds a majority overall. 
Researchers continue to debate which plural markers are productive, and in which circumstances.

The dispute has historically centered on the 
infrequent class \s, 
which, despite its rarity, occurs across a wide range of linguistic environments. Examples include proper names (e.g.\ \textit{der Bader $\rightarrow$ die Bader} `the barber $\rightarrow$ the barbers' but \textit{meine Freunden, die Baders} `my friends, the Barbers'), acronyms, and truncated and quoted nouns (e.g.\ \textit{der Asi $\rightarrow$ die Asis}, short for \textit{Asozialer} `antisocial person'). In addition, \mbox{/-s/} tends to be the plural class for recent borrowings from other languages, and children reportedly extend \s\ to novel nouns \citep{clahsenRegularIrregularInflection1992}. For these reasons, M95 argue that \s\ is the \textbf{default} plural: it applies in a range of heterogeneous \textbf{elsewhere} conditions which do not define a cohesive similarity space, serving as the ``emergency" plural form when other markers do not seem to fit. They further assert that, as the default form, \s\ is also the \textit{only} regular plural form, in the sense that it ``applies not to particular sets of stored items or to their frequent patterns, but to any item whatsoever" (\citeyear[192]{marcusGermanInflectionException1995}). Under this \textbf{minority-default} analysis, other German plural classes may be productive, but in a limited sense --- they can only extend to novel inputs which are similar in some respect to existing class members, while infrequent \s\ can apply to any noun regardless of its form \citep{clahsenLexicalEntriesRules1999}. M95 claim that this behavior should be particularly difficult for connectionist, i.e. neural, models to learn: \s\ cannot be generalized based on its frequency, as it is rare, and it cannot be generalized based on similar inputs, as it applies to heterogeneous, unfamiliar inputs.

Other researchers have challenged the minority-default account with evidence of regular, productive behavior from the two more common suffixes \e\ and \en. \en\ is argued to be the default class for feminine nouns and nouns ending with the weak vowel schwa \citep{wiesePhonologyGerman1996, dresslerWhyCollapseMorphological1999}, and children have also been found to overgeneralize \en\ \citep{kopckeAcquisitionPluralMarking1998}. \citet[][1025]{indefreyProblemsLexicalStatus1999} argues that \en\ and \e\ are ``regular and productive allomorphs with gender-dependent application domains'', noting that \e\ and \en\ are extended in elsewhere conditions where \s\ is blocked for phonological reasons, such as letters (\textit{die ``X"e}) and acronyms (\textit{die MAZen, Magnetaufzeichnungen}, `magnetic recordings'). \citet{bybeeRegularMorphologyLexicon1995} argues that, while \s\ does act as the default plural, it is still less productive than other plural classes due to its low type frequency.

\paragraph{Wug testing for German plurals} To assess whether German speakers treat \s\ as a productive default for novel words, M95 developed a list of 24 monosyllabic nonce nouns for wug testing. The stimuli represented two phonological classes: `familiar' or Rhyme words, which rhymed with one or more existing words in German (e.g. \textit{Bral}, rhyming with \textit{Fall}; \textit{Spert}, rhyming with \textit{Wert}), and `unfamiliar' or Non-Rhyme words (e.g. \textit{Plaupf}, \textit{Fn\"ohk}), which were constructed using rare but phonotactically valid phone sequences.  They hypothesized that Non-Rhymes, as phonologically atypical words, should be more likely to take the \s\ plural.
M95 conducted a rating study in which stimuli were presented across three different sentence contexts. If the word \textit{Bral} was presented in the ``root" condition, subjects would rate a set of sentences where the nonce word referred to some object: \textit{Die gr\"unen BRAL sind billiger} (``The green brals are cheaper"), \textit{Die gr\"unen BRALE \ldots}, \textit{Die gr\"unen BR\"ALE \ldots}, etc.; whereas in the ``name" condition, the nonce word would refer to people: \textit{Die BRAL sind ein bi{\ss}chen komisch} (``The Brals [family name] are a bit weird"), \textit{Die BRALEN \ldots}, \textit{Die BRALS \ldots}, etc. With data from 48 participants, \s\ was the top-rated plural form for 2 out of 12 rhyme words, and 7 out of 12 non-rhyme words; while \e\ was rated highest overall, \s\ was the only marker favored more for non-rhymes. \citet{clahsenDualNatureLanguage1999} cites this asymmetry as crucial evidence for \s\ as the only default plural form, at least with respect to these stimuli.

These results, however, have been called into question. \citet[][henceforth Z\&L]{zaretskyNoMatterHow2016} conducted a large-scale follow-up study with 585 participants, using the same nonce words but a different task: instead of rating the plural forms within a sentence context, subjects were presented with the noun in isolation (e.g. \textit{Der Bral}) and asked to produce its plural form.\footnote{Z\&L's data is unfortunately not freely available.} They found a much lower preference for \s\ than expected based on M95's results, and a significant effect for feminine (\textit{die}) versus non-feminine (\textit{der, das}) grammatical gender, where M95 reported no effect of gender. The authors conclude from their data that \en, \e, and \s\ are all productive in German, and also speculate that task differences (production versus rating) could account for the discrepancy between the two studies. 

\subsection{Data collection} \label{stim}

\paragraph{Motivation} 

Although M95 published average rating data for each word in the appendix to their paper, we felt it necessary to collect our own data. Z\&L's findings suggest that the M95 \s\ effect might reflect task artefacts: speaker behavior could differ for production and rating tasks, and with and without sentential context for the nonce words. 
We seek to evaluate K\&C's performance claims for ED models, 
which were based on speaker production probabilities rather than ratings. To do so, we need speaker data which closely parallels the model task: given a noun in isolation, produce its plural inflected form. We collect production data, and also ratings, to see whether speaker behavior is consistent across tasks.

Another issue raised by Z\&L's findings is the role of grammatical gender. Although Z\&L reported significant gender effects,
M95 did not: their reported rating averages combine all gender presentations (e.g. \textit{Der Bral, Die Bral, Das Bral}). Previous experiments have found neural models of German plurals to be sensitive to grammatical gender \citep{goebel2000recurrent}; therefore, the stimuli presented to speakers should be consistent with model inputs to enable valid comparison. For simplicity, we opted to select one grammatical gender for presentation: neuter, or \textit{Das}. Based on similar experimentation by
\citet{kopckeSchemasGermanPlural1988}, 
speakers do not have a strong majority class preference for neuter monosyllablic nouns, hence this environment may be the most challenging for a neural model to learn. For this reason, we present all stimuli as neuter to study participants.

\paragraph{Method} 
The current study uses the same Rhyme and Non-Rhyme stimuli from M95's original experiment. We collected both production and rating data on plural inflection for the 24 M95 nonce nouns through an online survey with 150 native German-speaking participants.
Survey respondents were first prompted to produce a plural-inflected form for each noun (i.e. filling in the blank: \textit{``Das Bral, Die \_\_\_\_"}).\footnote{The article \textit{das} indicates singular number, neuter gender; as all nouns were presented in neuter gender (see preceding discussion), all nouns were preceded by \textit{das}. \textit{Die} here indicates plural number, so the following noun will be pluralized.} After producing plural forms for all nouns, they were prompted to rate the acceptability of each potential plural form for each noun on a 1-5 Likert scale, where 5 means most acceptable. For example, a participant would see \textit{Das Bral}, and then give an acceptability rating for each of the following plural forms: \textit{Bral, Br\"al, Brale, Br\"ale, Bralen, Braler, Br\"aler, Brals}.
For details of the survey design, please see Appendix A.

\begin{table}[]
\begin{center}
         \begin{tabular}{ l l | l l | l}

         Plural & & Prod \% & N & Rating (SE) \\ 
         \hline
         \hline
         
         \multirow{2}{2em}{\mbox{/-e/}} & R & \textbf{45.3} & 815 & 3.53 (.021) \\ 
            & NR & \textbf{44.7} & 805 & 3.51 (.024) \\
        \hline    
         \multirow{2}{2em}{\mbox{/-(e)n/}} & R & 25.0 & 450 & \textbf{3.73} (.026) \\ 
            & NR & 34.7 & 624 & \textbf{3.84} (.025) \\
    \hline
         \multirow{2}{2em}{\mbox{/-er/}} & R & 17.4 & 314 & 3.08 (.022) \\ 
            & NR & 6.7 & 120 & 3.06 (.024) \\
    \hline
         \multirow{2}{2em}{\mbox{/-s/}} & R & 4.2 & 75 & 2.39 (.027) \\ 
            & NR & 6.4 & 116 & 2.52 (.028) \\
          \hline  
          \multirow{2}{2em}{\mbox{\z}} & R & 2.7 & 48 & 2.24 (.020) \\ 
            & NR & 2.7 & 48 & 2.38 (.024) \\
        \hline   
         \multirow{2}{2em}{other} & R & 5.4 & 98 &  \\ 
            & NR & 4.8 & 87 &  \\ 
         \hline
         \hline
         \multirow{2}{2em}{overall} & R &  & 1800 & 2.99 (.011) \\ 
            & NR &  & 1800 & 3.04 (.012) \\ 
        \end{tabular}
\caption{Survey results. Production reported as percentages out of all Rhymes (R) and Non-Rhymes (NR); ratings are averages over a 1 (worst) -- 5 (best) scale, with standard errors in parentheses. Highest numbers in each category are bolded.} 
\label{table:survey}
\end{center}
\end{table}

\subsection{Results} 

Our study results are shown in Table \ref{table:survey}. The production data collected in our survey appears broadly consistent with the distribution observed by Z\&L and \citeauthor{kopckeSchemasGermanPlural1988}: \e\ is favored in production, followed by \en. The rhyme vs non-rhyme comparison is also consistent with Z\&L's results. \s\ is produced more for Non-Rhymes than for Rhymes, as emphasized by \citet{clahsenDualNatureLanguage1999}; however, \en\ also shows the same directional preference, and at a much higher frequency.

Our rating results diverge from production results in some ways --- for example, \en\ is favored instead of \e\ --- and are consistent in others: both \s\ and \en\ are rated higher for Non-Rhymes compared to Rhymes. 
The low ratings for \s\ conflict with M95's findings, and suggest that presentation in sentence context is an important methodological difference from presentation in isolation.
For example, family surnames obligatorily take \s\ in German, so it's possible that exposure to surnames in the ``name" context primed subjects in the M95 rating study to find \s\ more acceptable generally, across conditions.\footnote{\citet{hahnGermanInflectionSingle2000} reanalyze the M95 ratings and find that \s\ is rated much higher for family surnames than other kinds of names within the ``name" condition (e.g. first names), reflecting the strong link between this category and the \s\ plural class.} In any case, our results demonstrate task effects: although \e\ is the most \textit{produced} plural form, \en\ obtains the highest \textit{ratings} from the same speakers.\footnote{Further analysis indicates that individual survey participants rated a plural form they did \textit{not} produce as better than the form they \textit{did} produce in fully one-third of cases.} 
We compare these results with the modeling study in Section \ref{discussion}, focusing on production data.

\section{Study 2: Encoder-Decoder inflection}

Our second study trains an encoder-decoder (ED) model on the task of German plural inflection, following the method of \citeauthor{kirovRecurrentNeuralNetworks2018} (K\&C). We then compare its predictions on the M95 stimuli to the behavior of participants in Study 1.

\subsection{Background}

\paragraph{Wug testing and computational models} 
Wug tests have also been used to evaluate how computational models generalize, although the appropriate method of comparison to speakers is still under debate. \citet{albrightRulesVsAnalogy2003} collected spoken productions and acceptability ratings of past tense inflections for English nonce verbs, comparing the prevalence of regular inflection (e.g. \textit{rife $\rightarrow$ rifed})) to one or two pre-selected irregular forms for each nonce verb (e.g. \textit{rife $\rightarrow$ rofe, riff}). 
They then evaluated two different computational models on their wug data, focusing on correlation between model scores and participant ratings to select a rule-based learner as the best-performing model. 
K\&C also tested their ED model on \citeauthor{albrightRulesVsAnalogy2003}' nonce words and 
evaluated performance using correlation with model scores; however, instead of the rating data, they focused on \textbf{production probabilities}: the percentage of speakers who produced each pre-selected irregular form. \citet{corkeryAreWeThere2019} call this methodology into question, noting that different random initializations of the ED model lead to 
highly variable rankings of the output forms, and thus to unstable correlation metrics. Instead, they correlate the speaker production probabilities to the aggregated predictions of models with different random seeds, treating each model instance as simulating a unique ``speaker". 
Our study follows the latter approach: we aggregate production probabilities over several model initializations and compare these results to the speaker production data.

\paragraph{Modeling German plurals} 
The same M95 stimuli used in our Study 1 have also been applied to wug test computational models. To date, no computational studies have reproduced the high \s\ preference reported for participants in the original rating study. \citet{hahnGermanInflectionSingle2000} framed the problem as a classification task, mapping noun inputs to 
their plural classes.
They trained a ``single-route" exemplar-based categorization model \citep{nosofskyExemplarbasedAccountsRelations1988} alongside a ``dual-route" version of the same model, which had an additional symbolic rule component to handle the \s\ class.
\citeauthor{hahnGermanInflectionSingle2000} also collected their own speaker productions of the M95 wug stimuli, 
and found that the single-route model showed a higher overall correlation to speaker production probabilities, relative to the dual-route model.
They did not explicitly compare model and speaker behavior on Rhymes versus Non-Rhymes, so we don't know whether the model learned speaker-like generalizations for phonologically atypical stimuli, or whether the model could achieve similar performance on the more challenging task of sequence prediction.

\citet{goebel2000recurrent} used a simple recurrent network \citep{elmanFindingStructureTime1990} for sequence prediction on the M95 wug stimuli.
The model did produce \s\ more often for Non-Rhymes than Rhymes, but as the overall production of \s\ was relatively low, the authors did not consider this evidence of default behavior. Instead, they find that the model learns to condition regular plural inflection on grammatical gender. For both Rhymes and Non-Rhymes, the model predicted \en\ when the input was preceded by the feminine article \textit{die}, and \e\ when the input began with masculine \textit{der}; neuter \textit{das} was not tested. \citeauthor{goebel2000recurrent} reanalyze the original M95 rating data and argue that its results are hypothetically\footnote{"Hypothetically" because M95 did not report results split by grammatical gender.} consistent with the model's behavior; they conclude that \s, \en, and \e\ are all regular plural classes in German, with the latter two conditioned on grammatical gender. These findings show the importance of controlling for grammatical gender in comparing speaker and model results.

\begin{table}[]
    \centering
    \begin{tabular}{l | l l l l}
 Plural  & \% All & Neut & M95 R & 1 Syll \\ [0.5ex] 
 \hline
 /-(e)n/ & \textbf{37.3} & 3.2 & 13.9 & 14.0 \\ 
 /-e/ & 34.4 & \textbf{51.9} & \textbf{72.6} & \textbf{66.5} \\
 \z & 19.2 & 21.5 & 0.5 & 1.4 \\
 /-er/ & 2.9 & 10.6 & 7.3 & 4.7 \\
 /-s/ & 4.0 & 7.7 & 3.1 & 12.5 \\ 
 other & 2.1 & 5.1 & 2.6 & .9\\ 
 \hline
 N &  11,243 & 2,606 & 642 & 570
    \end{tabular}
    \caption{\small{Distribution (percentages) of plural class for 1) nouns overall, 2) only neuter nouns, 3) nouns rhyming with M95 stimuli, 4) one-syllable nouns from Unimorph German dataset \citep{kirovVerylargeScaleParsing2016}}.}
    \label{tab:um}
\end{table}

\subsection{Method}

\paragraph{Overview} 
We model German number inflection using the sequence-to-sequence Encoder-Decoder architecture \citep{sutskever_sequence_2014}. This comprises a recurrent neural network (RNN) which reads in an input sequence and \textbf{encodes} it into a fixed-length vector representation, and another RNN which incrementally \textbf{decodes} that representation into an output sequence. Following \citet{kannSingleModelEncoderDecoderExplicit2016}, our decoder uses neural attention \citep{bahdanauNeuralMachineTranslation2015}.   

For our task of morphological transduction, the ED model takes character-level representations of German nouns in their singular form as inputs (e.g. \textsc{$\langle m \rangle$ h u n d $\langle eos \rangle$}), and learns to produce the noun's inflected plural form (e.g. \mbox{\textsc{h u n d e $\langle eos \rangle$}}). Each character sequence starts with $\langle m \rangle, \langle f \rangle$, or $\langle n \rangle$, to indicate grammatical gender. Unlike English, the phonological-orthographic mapping is straightforward in German, 
so we can use a written corpus for model training.
We keep a held-out dev set for hyperparameter selection, and a held-out test set to asses the model's accuracy in generalizing to unseen German nouns. 
In addition, the 24 M95 nouns were used for comparison with speaker behavior. They were presented to the model as neuter gender, consistent with Study 1.

\paragraph{Corpus}  We trained all models on the UniMorph German data set\footnote{\url{https://github.com/unimorph/deu}} \citep{kirovVerylargeScaleParsing2016, sylak-glassmanLanguageIndependentFeatureSchema2015}, which provides the singular and plural forms of 11,243 nouns. Only nominative case forms were used. Grammatical gender was obtained by merging the Unimorph dataset with a more recent Wiktionary scrape containing this feature.\footnote{\url{https://github.com/gambolputty/german-nouns/} To ensure our results were not limited by the small size of the UniMorph dataset, we also trained the model on this larger dataset, including about 65,000 nouns. As the outcome was consistent with our findings here, we report results from the smaller model.} Table \ref{tab:um} gives the distribution of plural suffixes for the UniMorph corpus overall, and for three relevant subsets: nouns with neuter gender, monosyllabic nouns (like the M95 stimuli), and nouns which were phonologically similar to the M95 stimuli, i.e.\ shared a rhyme.
The number of items in the train, dev, and test splits is shown (in parentheses) in Table \ref{tab:mod_acc}. 

\paragraph{Implementation} Following K\&C and \citet{corkeryAreWeThere2019}, our model is implemented using OpenNMT \citep{kleinOpenNMTNeuralMachine2018} with their reported hyperparameters \citep[after][]{kannSingleModelEncoderDecoderExplicit2016}: 2 LSTM encoder layers and 2 LSTM decoder layers, 
300-dimensional character embeddings in the encoder, and 100-dimensional hidden layers in both encoder and decoder; Adadelta optimization 
for training with a batch size of 20 and inter-layer dropout rate of 0.3; and a beam size of 12 for decoding during evaluation. 

Since \citet{corkeryAreWeThere2019} found the ED model to be highly sensitive to initialization, we trained multiple simulations with the same architecture, varying only the random seed. Reported results combine predictions from 25 separate random initializations. The one hyperparameter we tuned was early stopping. Best performance on the validation set was achieved at 10 epochs, which was sufficient to memorize the training data.

\begin{table}[]
    \centering
    \begin{tabular}{l l l}
Train & Dev & Test \\ [0.5ex] 
 \hline
 99.9\%  (8694) & 92.1\% (1229) & 88.8\% (1320) 
    \end{tabular}
    \caption{\small{Model accuracy (N) by UniMorph corpus split, averaged over 25 random initializations.}}
    \label{tab:mod_acc}
\end{table}

\begin{table}[]
    \centering
    \begin{tabular}{l | l l l | l l l}
  & Test &  & & M95  \\ [0.5ex] 
   & Prec. & Rec. & F1 
                        & \%R & \%NR & $\rho$\\
 \hline
 /-(e)n/ & .95 & .95 & .95 
                            & 6.3 & 3.3 & .28 \\ 

 /-e/ & .86 & .89 & .87 
                        & \textbf{68.3} & \textbf{91.7} & .13 \\

 \z & .96 & .91 & .92 
                            & 0 & 0 & \\

 /-er/ & .83 & .85 & .84 
                        & 21.7 & 2.7 & .05 \\
 /-s/ & .64 & .56 & .60 
                        & 3.7 & 2.3 & .33 \\ 
 other & .37 & .48 & .42 
                        & 0 & 0 & \\ 
    \end{tabular}
    \caption{\small{Model results by plural suffix for: (left) test set performance (averaged over plural seed); (right) production percentages for rhyme (R) and non-rhyme (NR) M95 stimuli, and correlation (Spearman's $\rho$) to speaker productions.}}
    \label{tab:mod_results}
\end{table}

\begin{figure*}
    \centering
    \includegraphics[width=6in]{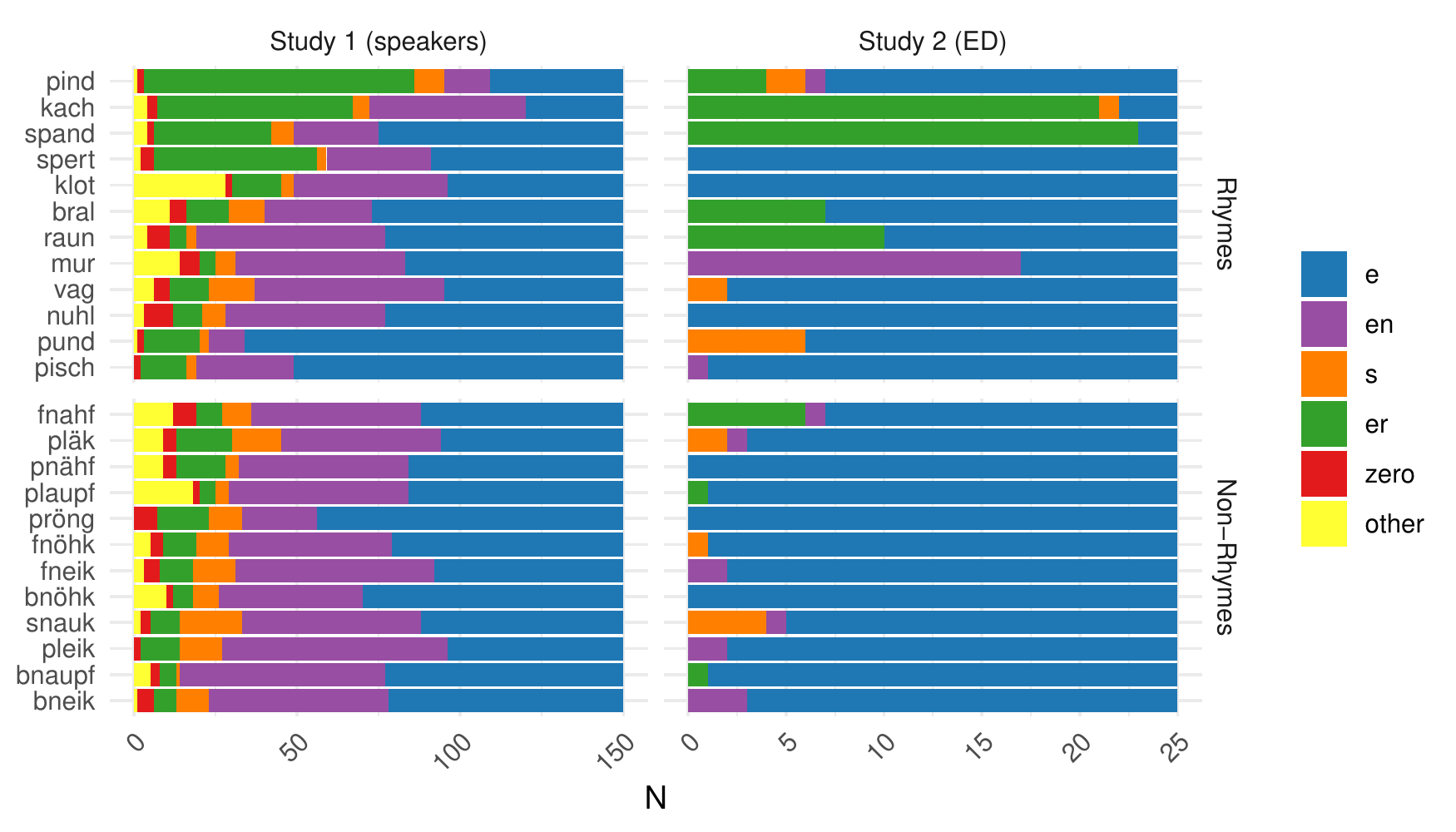}
    \caption{Plural class productions by item.}
    \label{fig:item_prod}
\end{figure*}

\paragraph{Results} The model achieves 88.8\% accuracy on the held-out test set (Table \ref{tab:mod_acc}). 
It performs best on \en, the most frequent class (Table \ref{tab:mod_results}).  
Unsurprisingly, the worst performance appears on the `other' category, which comprises the long tail of idiosyncratic forms which must be memorized (e.g. Latinate plurals \textit{Abstraktum $\rightarrow$ Abstrakta} or other borrowings \textit{Zaddik $\rightarrow$ Zaddikim}). 
In keeping with the findings of
\citet{hahnGermanInflectionSingle2000}, \s\ is the plural suffix with the worst generalization performance; this cannot be attributed to low frequency alone (c.f. Table \ref{tab:um}), as the model does much better on the similarly rare suffix \er\ . 

We use the M95 stimuli to compare model predictions to speaker data from Study 1.
The model shows an overwhelming preference for \e\ on these words (Table \ref{tab:mod_results}); roughly 80\% of its productions are \e, relative to 45\% of speaker productions (Figure \ref{fig:item_prod}). In contrast, the model rarely predicts \en, which speakers use 30\% of the time. The model's treatment of Rhymes and Non-Rhymes is even farther off the mark: where speakers use \en\ and \s\ \textit{more} for Non-Rhymes relative to Rhymes, the ED model uses them \textit{less}, producing \e\ for over 90\% of Non-Rhymes. Following K\&C and \citet{corkeryAreWeThere2019}, we calculate the Spearman rank correlation coefficient (Spearman's $\rho$) between model and speaker production probabilities within inflectional categories rather than across categories.\footnote{For the English analyses in the prior works, this means calculating separate correlations for regular and irregular forms.}
This means that, for each potential plural suffix, we compare speaker and model productions for that suffix on each individual M95 word.
Table \ref{tab:mod_results} reports the correlation for each suffix. None show a statistically significant difference from the null hypothesis of no correlation.

 \begin{figure*}[h!]
    \centering
    \includegraphics[width=6in]{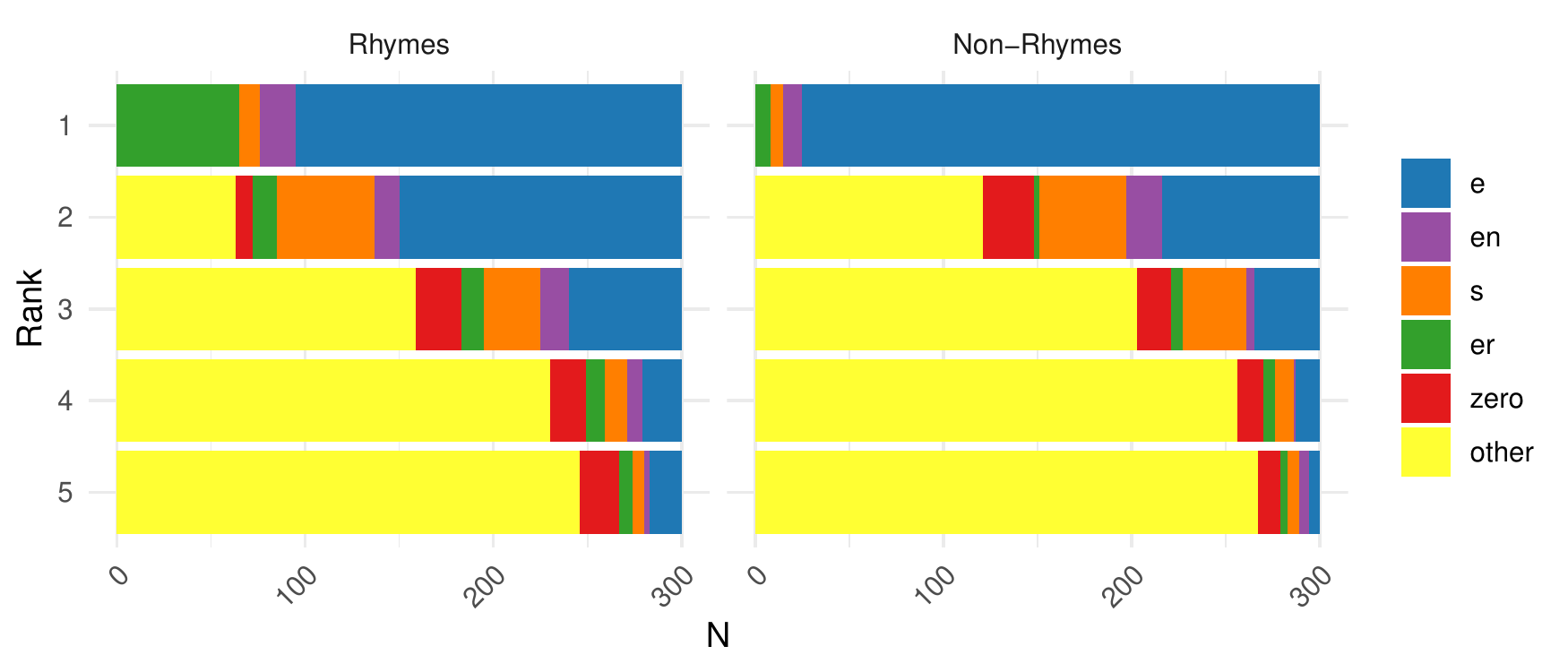}
    \caption{Distribution of plural classes by rank in ED model output.}
    \label{fig:item_rank}
\end{figure*}

Figure \ref{fig:item_rank} shows the distribution of plural classes in the top 5 most likely forms predicted by the model for each M95 word. 
While all of the model's top-ranked predictions are well-formed outputs in the sense that they conform to one of the main German plural classes, the lower-ranked predictions are rapidly dominated by ``other" forms which do not cohere to standard plural production. An example from one model instance: the Rhyme input \textit{Spert} had as its top five predictions \textit{Sperte, Spelte, Spente, Sperten,} and \textit{Fspern}; the Non-Rhyme input \textit{Bneik} had \textit{Bneiken, Bneiks, Bneikke, Bneikz,} and \textit{Bneikme}. 
\citet{corkeryAreWeThere2019} observed instability in the ranking of irregular forms in ED models trained on the English past tense; however, English irregular forms are very diverse, which makes it difficult to draw broad conclusions about the plausibility of lower-ranked forms in the model's output. In contrast, the five main plural suffixes for German cover 98\% of the nouns in the UniMorph dataset, and 95\% of speaker productions on M95 stimuli in Study 1. The predominance of ill-formed plurals in lower-ranked predictions\footnote{Interestingly, while less frequent classes such as \s\ and \z\ appear more often in the model's lower-ranked outputs, the class \en\ is almost never predicted --- despite being the second most frequent class in speaker data productions.} suggests ED model scores may not be 
cognitively plausible analogues to speaker behavior; if they were, we would expect forms with standard plural inflections to receive consistently high rankings.

\section{Discussion} \label{discussion}

The current study asks whether modern Encoder-Decoder neural models learn the full set of correct generalizations --- that is, human-like behavior --- with respect to German number inflection, which requires the learner to generalize non-majority inflectional classes. 
The short answer is no: our model learns \textit{part} of that set. In particular, it correctly identifies \e\ as the `best' plural class for this context. \e\ is the most frequent class in the training data for similar inputs (neuter gender, monosyllabic, phonologically close to M95; c.f. Table \ref{tab:um}), and it is also the plural suffix most frequently produced by speakers (Table \ref{table:survey}). Like all plural classes, \e\ does not characterize a majority of German nouns overall (Table \ref{table:de_plurals}), so the model has technically learned to generalize a minority class in its appropriate context. Nonetheless, it does not reproduce the behavior of survey participants in response to the same stimuli, which shows a more variable distribution over plural classes and different generalization patterns for Non-Rhymes relative to Rhymes.

This outcome is not surprising when one considers that the model is trained to produce one correct form rather than a distribution over plausible forms; however, this is exactly the task faced by human language learners as well. All the models of morphology discussed here assume that exposure to correct forms alone should suffice for learning speaker-like behavior. \citet[][3872, fn. 4]{corkeryAreWeThere2019} note that training on single target forms produces highly skewed ED model scores, with a great deal of probability mass on the top-ranked form and instability in lower rankings, but that training on a distribution would not be a cognitively plausible alternative. 
However, it could be the case that German speakers do regularly encounter variable realizations of plural forms. \citeauthor{kopckeSchemasGermanPlural1988} observes that German plural inflection shows regional variation, for example northern speakers using \s\ (\textit{die M\"adels} `girls') where southern dialects prefer \en\ (\textit{die M\"adeln}). Incorporating dialect-informed variability into training might be one way to encourage neural models toward speaker-like generalization.\footnote{Like previous studies on these stimuli, our Study 1 did not collect data on speakers' dialect background; we are addressing this issue in follow-up research. We note that Study 1 began with an onboarding task prompting speakers to inflect existing nouns in Modern High German, which hopefully primed use of the standard variety for the following tasks.}

Parallel issues arise for model evaluation: how should we evaluate models of production when the target output is a distribution? On simplified versions of the task, such as classification \citep{hahnGermanInflectionSingle2000}, the output distribution is constrained within a space of plausible forms, but sequence-to-sequence models deal with the open-ended domain of all possible strings. For encoder-decoders, the likelihood scores produced during beam-search decoding offer an intuitive option, and K\&C use these scores to evaluate their model with respect to \citeauthor{albrightRulesVsAnalogy2003}' wug data; however, \citet{corkeryAreWeThere2019} demonstrate that these model scores are not a suitable metric for that comparison. Other recent research has highlighted the limitations of both beam search and model scores globally in neural sequence-to-sequence models \citep{stahlbergNMTSearchErrors2019}. Our results provide further evidence that lower-ranked ED predictions do not reflect cognitively plausible distributions: they contain many ill-formed outputs, and omit inflectional classes such as \en, which is prevalent in speaker productions. An alternative to model scores is to treat each randomly initialized instance of a model as an individual, and compare aggregate productions with speaker data \citep{goebel2000recurrent, corkeryAreWeThere2019}. For our experiments, this did not produce the distribution observed in the speaker data. The discrepancy between speaker production and rating preferences poses another challenge, as it's not clear how the ED model might represent these different task modalities.

Beside variability, the other key discrepancy between speaker and ED behavior is the treatment of Non-Rhyme words. If German has a default plural class, it should be realized more often on these phonologically atypical stimuli than the more familiar Rhyme words. Speakers in Study 1 use \s\ and \en\  more for Non-Rhymes than for Rhymes. These results are consistent with earlier studies: M95 found that \s\ was the only plural form to receive higher average ratings for Non-Rhymes compared to Rhymes, and Z\&L found that speakers produced both \en\ and \s\ more often for Non-Rhymes.
In contrast, the ED model appears to treat \e\ as a default, producing \e\ inflections for under 70\% of Rhymes but over 90\% of Non-Rhyme inputs.
This asymmetry suggests that the model has not induced the full set of correct generalizations for German plural inflection --- it has not recognized which plural classes are more productive for phonologically atypical nouns. In fact, the model's preference for \e, the most frequent (if non-majority) suffix, is the behavior anticipated by M95: ``frequency in the input to a pattern associator causes a greater tendency to generalize" (\citeyear[][215]{marcusGermanInflectionException1995}). It seems that the productivity of less frequent inflectional classes continues to challenge neural models and limit their
cognitive application.

\section{Conclusions}

German number inflection has been claimed to have distributional properties which make it difficult for neural networks to model. Our experimental speaker data does not necessarily support all of these claims; in particular, \s\ does not appear to be the only plural suffix which speakers treat as a `default' for phonologically unfamiliar words, as the more frequent marker \en\ shows similar trends. Nonetheless, the German plural system continues to challenge ED architectures. Our neural model struggles to accurately predict the distribution of \s\ for existing German nouns. On novel nouns, it generalizes the contextually
most frequent plural marker \e; its predictions are less variable than speaker productions, and show different patterns of response to words which are phonologically typical (Rhymes) as opposed to atypical (Non-Rhymes). Regardless of the minority-default question, it seems that ED models do not necessarily function as good cognitive approximations for inflectional systems like German number, in which no class holds the majority.

\section*{Acknowledgments}

The authors thank Yevgen Matusevych, Maria Corkery, Timothy O'Donnell, the Agora reading group at Edinburgh, and the ACL reviewers for helpful feedback. This work was supported in part by the EPSRC Centre for Doctoral Training in Data Science, funded by the UK Engineering and Physical Sciences Research Council (grant EP/L016427/1) and the University of Edinburgh. This work was also supported by a James S McDonnell Foundation Scholar Award (\#220020374) to the second author.

\bibliography{de_pl}
\bibliographystyle{acl_natbib}

\clearpage

\appendix

\section{Study design}

\subsection{Stimuli}

Table \ref{tab:stim} provides the complete list of nouns used in the experiment.

\begin{table}[h]
    \centering
    \begin{tabular}{c|c}
        Rhymes & Non-rhymes \\
        \hline
        Bral & Bnaupf \\
         Kach & Bneik \\
         Klot & Bn\"ohk \\
         Mur & Fnahf \\
         Nuhl & Fneik \\
         Pind & Fn\"ohk \\
         Pisch & Plaupf \\
         Pund & Pleik \\
         Raun & Pl\"ak \\
         Spand & Pn\"ahf \\
         Spert & Pr\"ong \\
         Vag & Snauk \\
    \end{tabular}
    \caption{Experimental stimuli \citep{marcusGermanInflectionException1995}}
    \label{tab:stim}
\end{table}

\subsection{Procedure}

We designed an online survey comprising three sections, in order of presentation: 1) an introductory production task with existing German words, 2) a nonce-word production task, and 3) a nonce-word rating task. For the introductory production task, eight existing German nouns were used, one from each of the eight plural classes under consideration. The goal of this section was to familiarize participants with the task of producing the plural, and avoid biasing them toward any particular plural marker by showing all eight options. We also hoped that inflecting nouns in Modern High German would encourage participants to approach the following tasks with the standard variety primed, thus reducing the possible effects of dialectal variation.
For the second and third sections, the production and rating tasks, the twenty-four M95 nonce words were presented. All stimuli were presented with neuter grammatical gender in the nominative case. In all tasks, each noun was preceded by the article \textit{Das}, indicating neuter gender and singular number, and each prompt for participant responses was preceded by \textit{Die...}, to indicate plural number. The eight existing nouns presented in the introductory production task were selected for neuter gender, so they followed this pattern as well.

We recruited 192 participants through the online survey platform Prolific\footnote{\url{http://www.prolific.com}}, using the site's demographic filters to target native German speakers. Participants were additionally asked about their age and exposure to languages other than German within the survey. Participants were shown the three tasks, introduction, production, and rating, in order, meaning that participants had to produce a plural form for all 24 nonce words before performing the rating task. For the production task, participants saw the noun on its own, preceded by \textit{Das}, e.g.\ \textit{Das Bral}.
Above the response box, the text \textit{Die...} appeared, to indicate that a plural form of the noun should be typed into the response box below the text.
For the rating task, participants were prompted to rate each potential plural on a Likert scale of \textit{Sehr gut} (`very good'; 5) to \textit{Sehr schlecht} (`very bad'; 1). 
After filtering out 42 respondents who failed a preliminary attention check, data from 150 participants was available for analysis. The cleaned, anonymized survey data will be published online along with this paper.

\end{document}